\documentclass[runningheads]{llncs}
\usepackage{graphicx}
\usepackage{subfigure}
\usepackage{caption}
\usepackage{makeidx}
\usepackage{amsmath,amssymb,theorem,bm,exscale}
\usepackage{textcomp,bm}
\usepackage[latin1]{inputenc}
\usepackage{latexsym}
\usepackage{placeins}
\usepackage{color}
\usepackage{lineno}
\usepackage{array}
\usepackage{multirow}
\usepackage{enumitem}
\usepackage[pagebackref=false,breaklinks=true,letterpaper=true,colorlinks,urlcolor=blue,citecolor=blue,linkcolor=blue,bookmarks=false]{hyperref}

\begin{document}
\title{Attentive CT Lesion Detection Using Deep Pyramid Inference with Multi-Scale Booster}
\titlerunning{CT Lesion Detection Using MSB}
% If the paper title is too long for the running head, you can set
% an abbreviated paper title here
%
\author{Qingbin Shao\inst{1,2\star} \and Lijun Gong\inst{1}\thanks{Q. Shao and L. Gong contribute equally and share the first authorship. L. Gong is the corresponding author. This work was done when Q. Shao was an intern in Tencent Youtu Lab. The source code and results are available at \url{https://github.com/shaoqb/multi_scale_booster.}} \and
Kai Ma\inst{1} \and Hualuo Liu\inst{2} \and Yefeng Zheng\inst{1}}
% index{Gong, Lijun}
% index{Shao, Qingbin}
% index{Ma, Kai}
% index{Liu, Hualuo}
% index{Zheng, Yefeng}

%
\authorrunning{Shao et al.}
% First names are abbreviated in the running head.
% If there are more than two authors, 'et al.' is used.
%
\institute{Tencent Youtu Lab, Shenzhen, China\\
\and Jilin University, Changchun, China\\
\email{lijungong@tencent.com}}
\maketitle              % typeset the header of the contribution
\begin{abstract}
Accurate lesion detection in computer tomography (CT) slices benefits pathologic organ analysis in the medical diagnosis process. More recently, it has been tackled as an object detection problem using the Convolutional Neural Networks (CNNs). Despite the achievements from off-the-shelf CNN models, the current detection accuracy is limited by the inability of CNNs on lesions at vastly different scales. In this paper, we propose a Multi-Scale Booster (MSB) with channel and spatial attention integrated into the backbone Feature Pyramid Network (FPN). In each pyramid level, the proposed MSB captures fine-grained scale variations by using Hierarchically Dilated Convolutions (HDC). Meanwhile, the proposed channel and spatial attention modules increase the network's capability of selecting relevant features response for lesion detection. Extensive experiments on the DeepLesion benchmark dataset demonstrate that the proposed method performs superiorly against state-of-the-art approaches.

\end{abstract}
\section{Introduction}
Automatically detecting lesions in CT slices is important to computer-aided detections/diagnosis (CADe/CADx). The identification and analysis of lesions in the clinic practice benefit the diagnosis of diseases at the early stage. The recent progress of the CADx mainly focuses on the visual recognition.
By using the Convolutional Neural Networks (CNNs), the automatic detection of lesions has reduced the workload of the manual examinations. These lesion detection approaches arise from the object detection frameworks such as Faster R-CNN \cite{ren-pami16-faster} and Feature Pyramid Network (FPN) \cite{lin-iccv17-fpn}, which typically employ a two-stage process. First, they draw a set of bounding box samples indicating the potential region-of-interest (ROI) on the feature maps of CT slices. Then, each sample is classified as either lesion or background by a binary classifier. The two-stage CNN based detection frameworks have been trained in an end-to-end fashion and achieved the state-of-the-art performance.
%The end-to-end learning has improved the detection accuracy into the state-of-the-art performance.

\renewcommand{\tabcolsep}{1pt}
\def\swone{0.22\linewidth}
\begin{figure}[t]
\centering
\small
\begin{tabular}{cccc}
    \vspace{-0.5mm}\includegraphics[width=\swone]{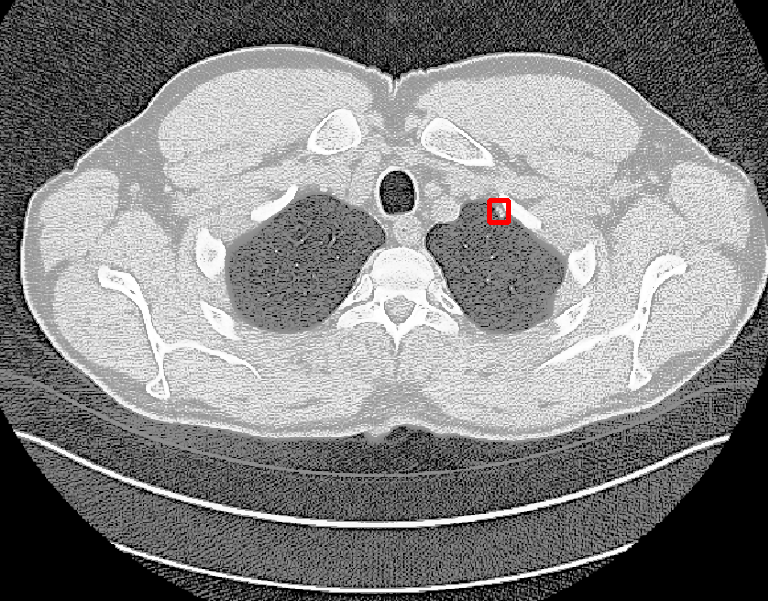}&
    \includegraphics[width=\swone]{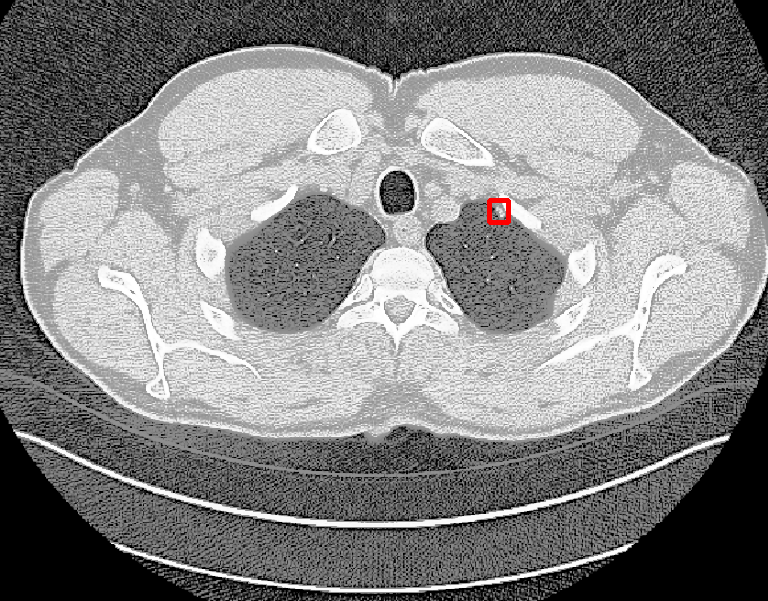} &
    \includegraphics[width=\swone]{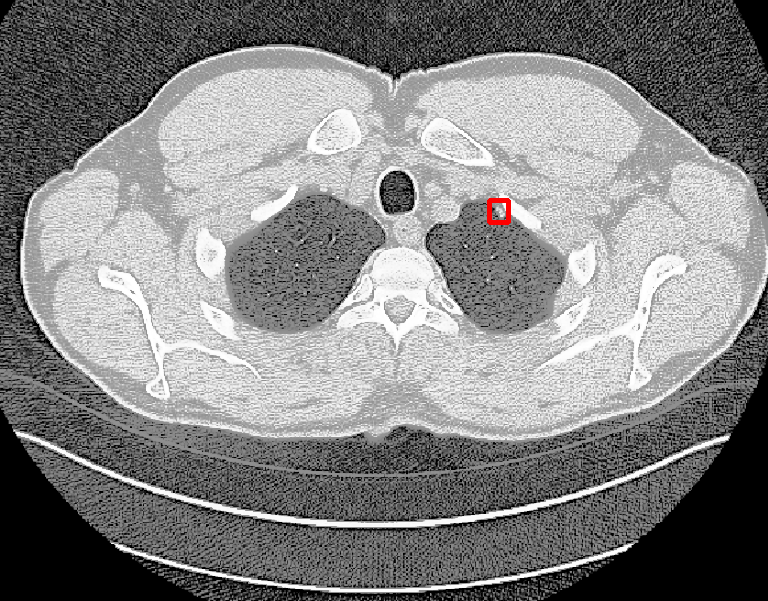} &
    \includegraphics[width=\swone]{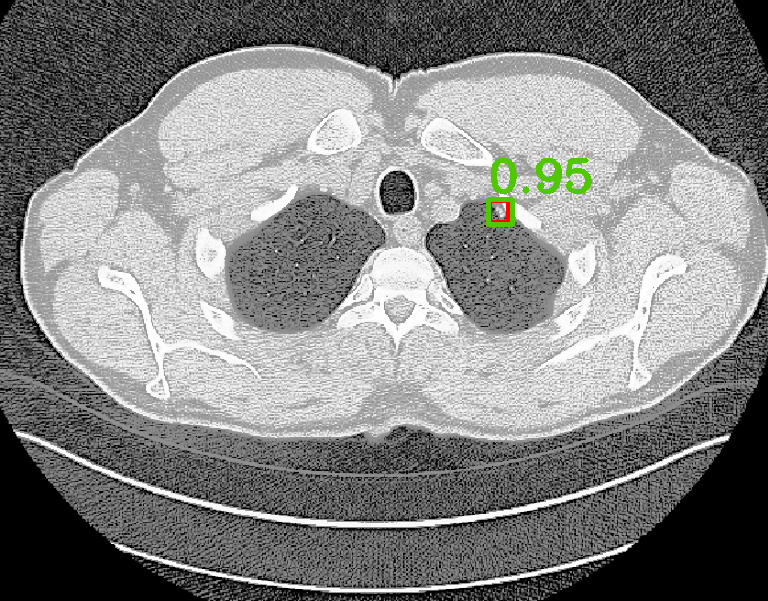}\\
    \vspace{-0.5mm}\includegraphics[width=\swone]{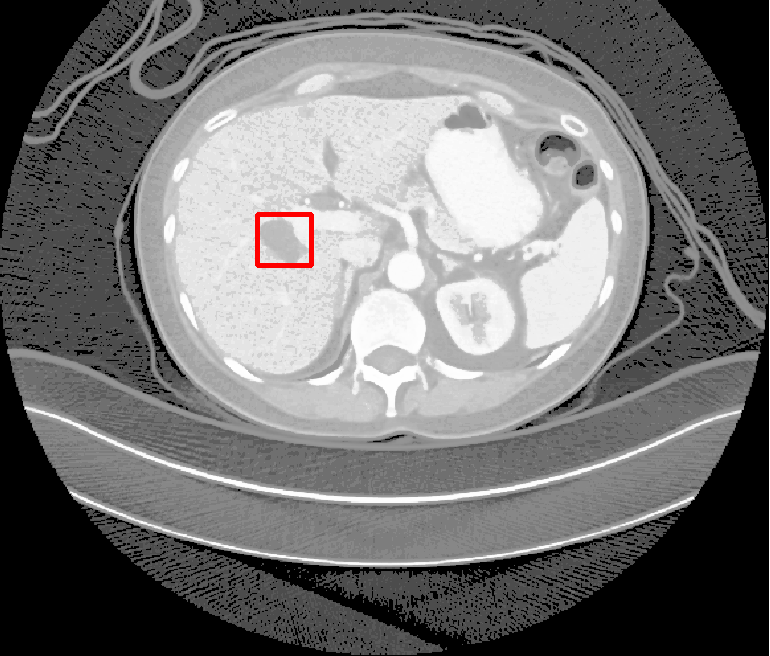}&
    \includegraphics[width=\swone]{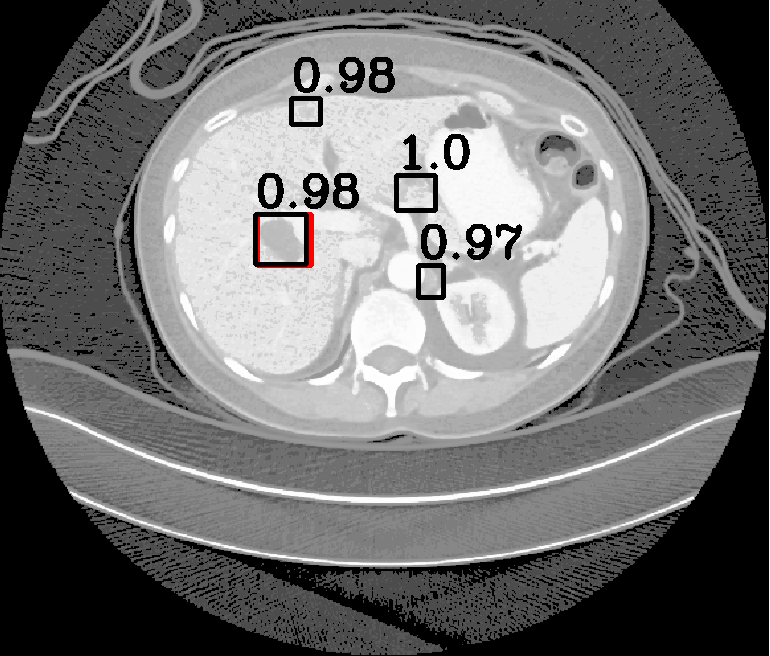} &
    \includegraphics[width=\swone]{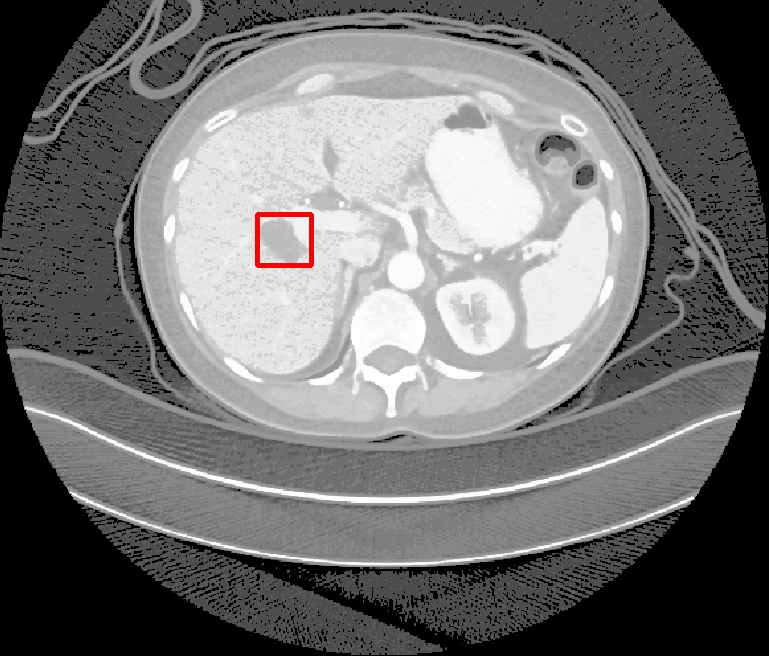} &
    \includegraphics[width=\swone]{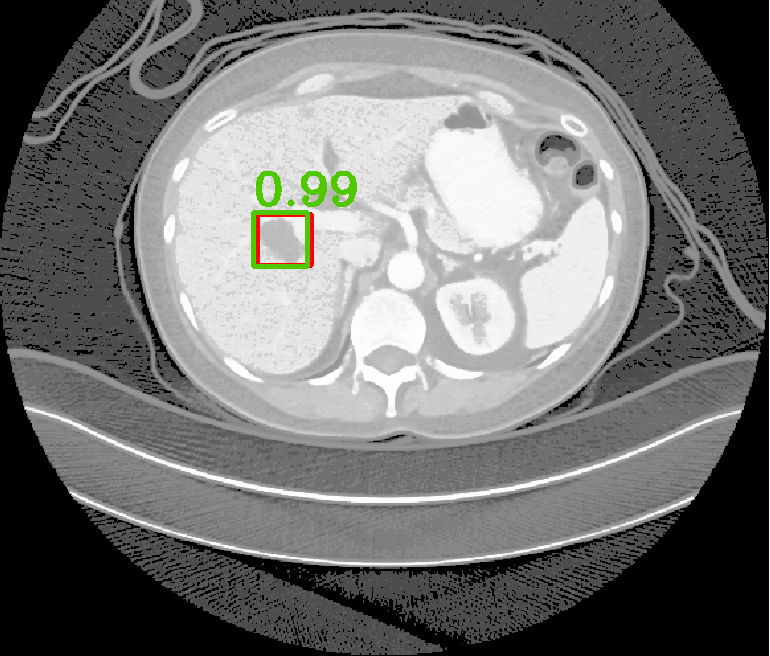}\\
    \vspace{-0.5mm}\includegraphics[width=\swone]{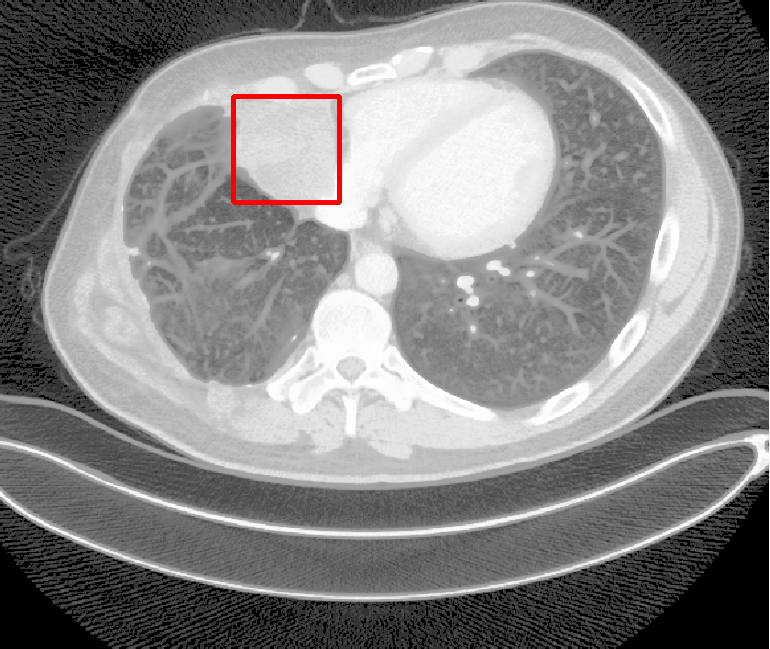} &
    \includegraphics[width=\swone]{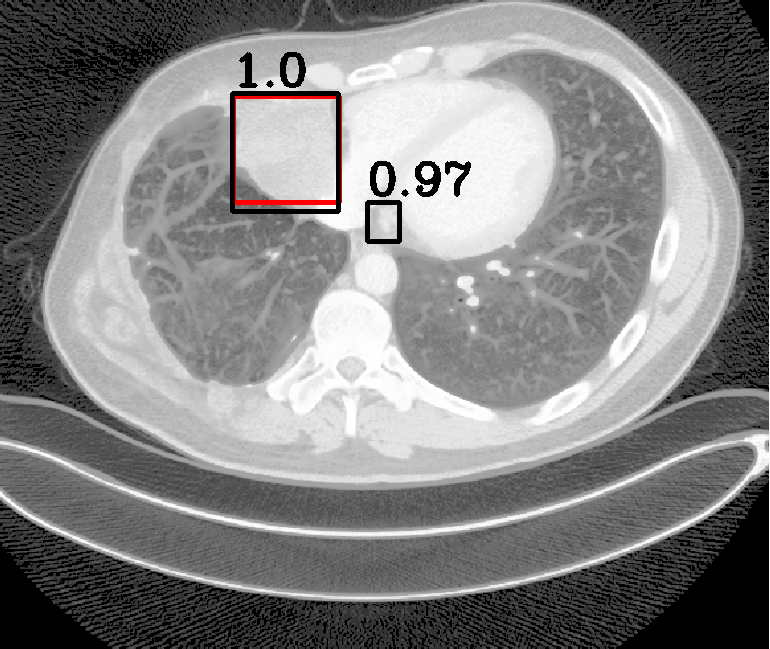} &
    \includegraphics[width=\swone]{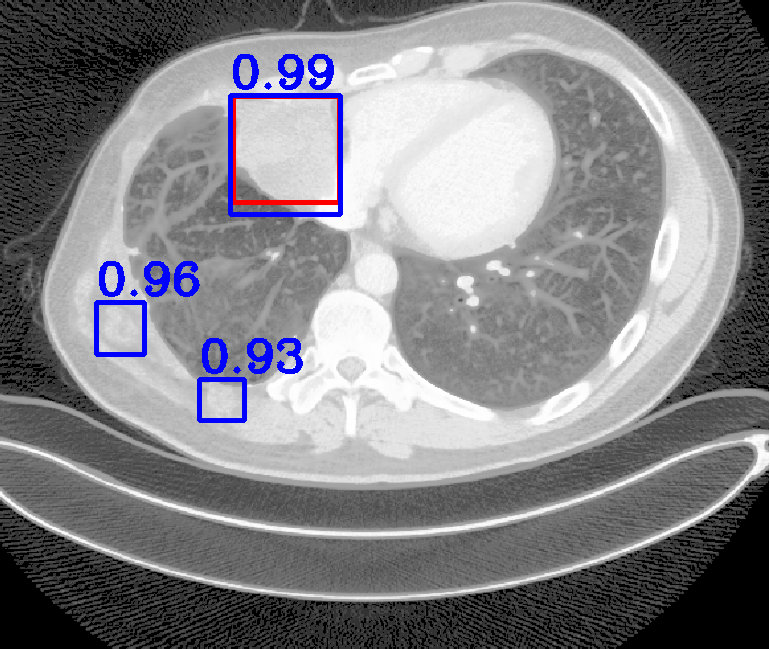} &
    \includegraphics[width=\swone]{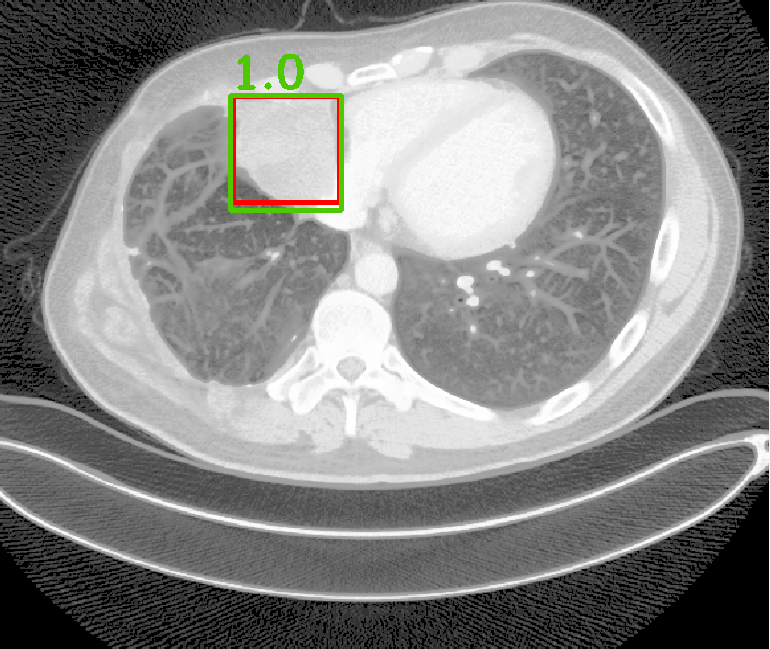}\\
     (a) Ground Truth & (b) Faster R-CNN & (c) FPN & (d) Proposed \\
\end{tabular}
\caption{Lesion detection results. The red bounding boxes represent ground truth annotations. The black, blue and green bounding boxes are the predicted results by Faster R-CNN \cite{ren-pami16-faster}, FPN \cite{lin-iccv17-fpn} and the proposed method, respectively.}
\label{fig:example}
%\vspace{-2mm}
\end{figure}

%Existing methods \cite{ding2017accurate,dou2017multilevel,liao2019evaluate,yan20183d} enrich the spatial information of CT scans by using 3D context representations.
To further improve the detection accuracy of CT data where blur and artifacts rarely exist \cite{song2017fast,song2018joint,roy2018concurrent}, several methods \cite{ding2017accurate,dou2017multilevel,liao2019evaluate,yan20183d} have been proposed to leverage the 3D spatial information. Ding et al. \cite{ding2017accurate} proposed a 3D-CNN classifier to refine the detection results of the pulmonary cancer from the 2D-CNN framework. Furthermore, Dou et al. \cite{dou2017multilevel} explored a 3D-CNN for false positive reduction in pulmonary nodules detections. On the other side, Liao et al. \cite{liao2019evaluate} extended the region proposal network (RPN) \cite{ren-pami16-faster} to 3D-RPN to generate 3D proposals. Although spatial representations extracted from 3D space improve the network performance on certain tasks, these methods suffer from tremendous memory and computational consumption. To tackle the computation efficiency problem, Yan et al. \cite{yan20183d} proposed a 3D context enhanced region-based CNN (3DCE) to produce 3D context from feature maps of 2D input images.
%Although spatial representations have been extracted in the 3D space, these methods suffer from tremendous memory and computational consumption.
It achieved similar performance to 3D-CNN while consuming the same speed of the traditional 2D-CNN, which deserves further improvement with more advanced networks.

In real-world scenarios, body lesions usually have arbitrary size. For instance, in the DeepLesion \cite{yan2017deeplesion} dataset, the lesion size ranges from 0.21 mm to 342.5 mm. Since most of the established CNNs are not robust to handle such spatial scale variations, they have unpredictable behavior in the varying cases. As shown in Fig. \ref{fig:example}, both Faster R-CNN and FPN fail to detect tiny lesions in the first row, while they produce small false positive lesions around the actual large lesion locations in the second and third rows.

In this paper, we propose a fine-grained lesion detection approach with a novel multi-scale attention mechanism. We use 2D FPN as the backbone to construct the feature pyramid in a relatively coarse scale. Within each level of the feature pyramid, we propose to use a Multi-Scale Booster (MSB) to facilitate lesion detection across fine-grained scales. Given the feature maps from one pyramid level, MSB first performs Hierarchically Dilated Convolution (HDC) that consists of several dilated convolution operations with different dilation rates \cite{yu-iclr15-dilatedConv}. The feature responses from HDC contain fine-grained information that is complementary to the original feature pyramid, which is achieved by extensive feature extraction in 2D space. The over-sampled feature responses are then concatenated and further exploited by channel-wise and spatial-wise attention. The channel attention module in MSB explores different lesion responses from the subchannels of the concatenated feature maps. The spatial attention module in MSB locates lesion response within each attentive channel. The channel-wise and spatial-wise attention modules enable the network to focus on particular lesion responses offered by the fine-grained features, while annealing the irrelevant and interference information. Thorough experiments demonstrate that MSB improves the deep pyramid prediction results and performs favorably against state-of-the-art approaches on the DeepLesion benchmarks.

\section{Proposed Method}
Fig. \ref{fig:pipeline} shows an overview of the pipeline. Our method uses a pre-trained FPN network to extract features from the input image at different pyramid levels. The extracted features are further processed by channel and spatial attention modules to capture fine-grained information to handle large spatial scale variations. The output of the MSB modules is used to make the final prediction at each pyramid feature map respectively.
%The proposed MSB uses the backbone FPN to detect lesions. Figure \ref{fig:pipeline} shows the overall pipeline and the architecture of MSB.
%We use MSB to supplement FPN for feature extraction and the output of MSB is used to make the final prediction at each pyramid feature map.

\def\swone{0.75\linewidth}
\begin{figure}[t]
  \centering
  \begin{tabular}{c}
  \includegraphics[width=\swone]{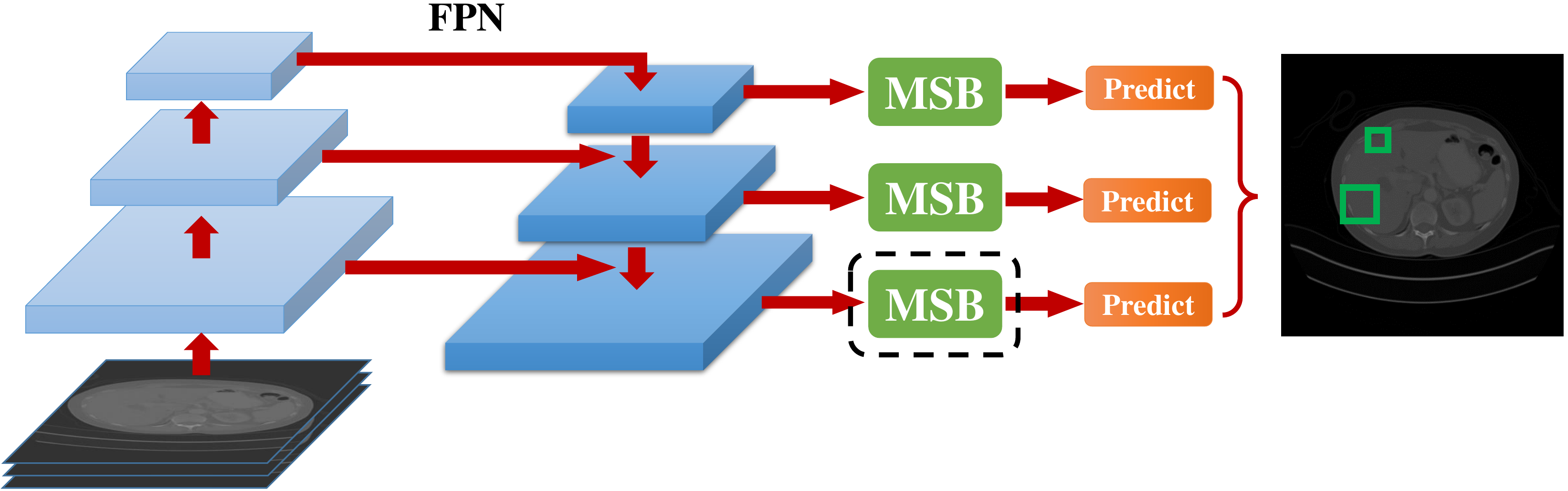}\\
  \includegraphics[width=\swone]{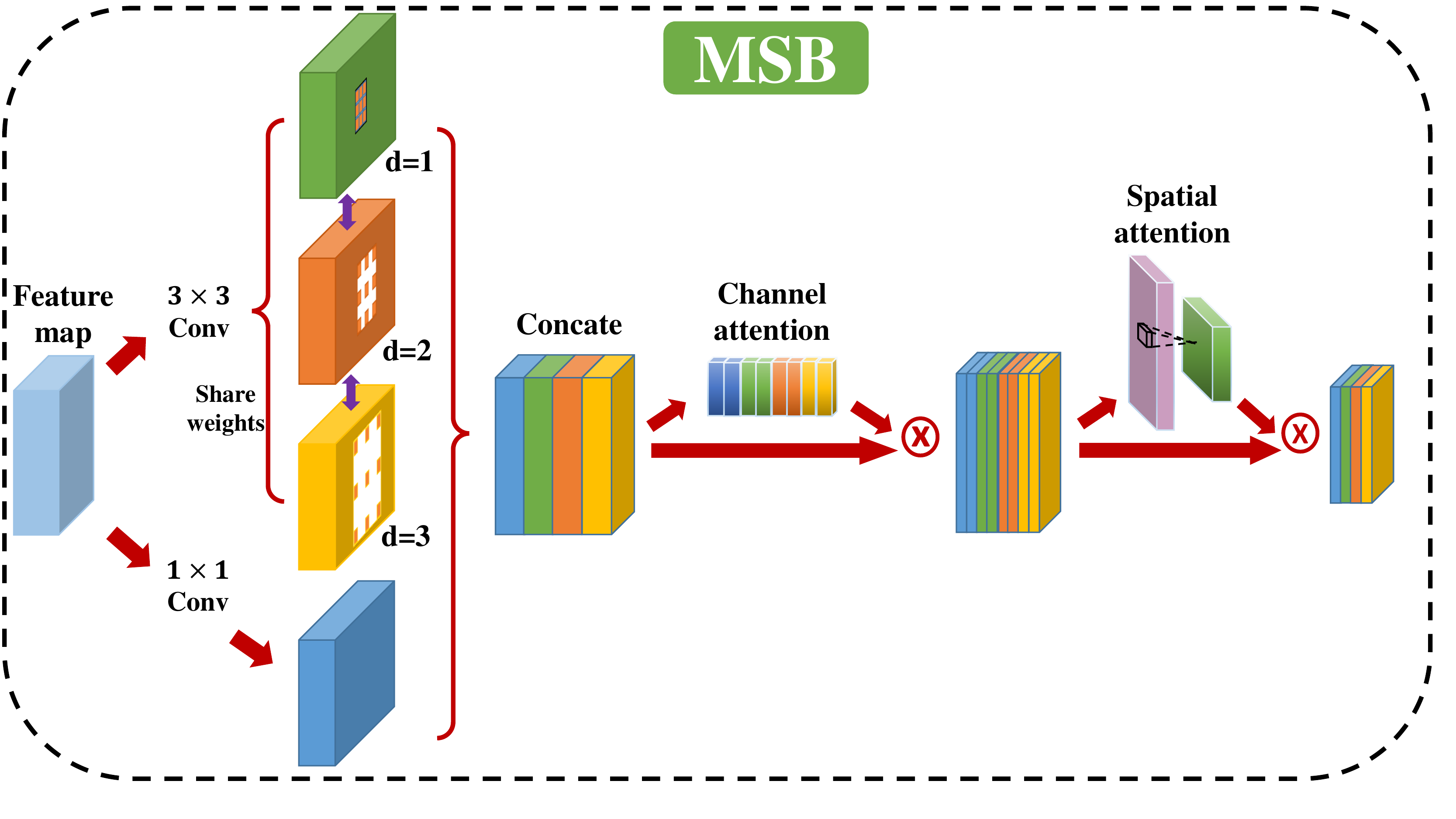}\\
  \end{tabular}
  \vspace{-3mm}
  \caption{Frameworks of the proposed approach. The detailed architecture of the Multi-Scale Booster (MSB) module is shown in the second row.}
  \label{fig:pipeline}
\end{figure}

\subsection{Revisiting FPN}
The FPN \cite{lin-iccv17-fpn} consists of three components for object detection: the bottom-up pathway, top-down pathway and skip connections in between. The bottom-up pathway computes feature maps at several different scales with a down-sampling factor of 2. We use $C_i^D$ to denote the feature maps at the $i$-th down-sampled pyramid. The $C_i^D$ has strides of $2^{i+1}$ pixels with respect to the input image. In the top-down pathway, the feature maps from the coarse levels are upsampled gradually to the finer resolutions with an up-sampling factor of 2. We denote the upsampled feature maps at the $i$-th upsampled pyramid as $C_i^U$ . The skip connections merge the downsampled and upsampled feature maps together at each pyramid level and the fused feature maps can be written as:
\begin{equation}
%\small
P_i=C_i^D \oplus C_i^U
\label{eq:fpn}
\end{equation}
where $\oplus$ is the element-wise addition operation. After generating the feature maps $P_i$, the potential objects are then detected at each feature pyramid level.

\subsection{Hierarchically Dilated Convolution}
%The dilated convolution captures multiple scales of reception fields by using different dilation rates.
The dilated convolution is commonly used to expand the reception fields without loss of the original resolution. In ASPP \cite{chen2017deeplab}, dilated convolution provided precise scale estimations for pixel-level semantic segmentation. Given the input feature map $C_i^D$, the dilated convolution can be written as:
\begin{equation}\label{eq:dc}
%\small
  y(x)=\sum_{k}C_i^D(x+r\cdot k)\cdot W(x)
\end{equation}
where $x$ is the location of the current pixel under processing; $k$ is the supporting pixels in the convolution process; $W$ is the filter weight; and $r$ is the dilation rate. The dilation rate corresponds to the stride that we use to sample input feature map $C_i^D$. We denote the dilated convolution in a general form as $\mathcal{D}_r(C_i^D)$ where $\mathcal{D}_r$ is the dilated convolution operator with dilation rate $r$.

The HDC performs multiple dilated convolutions with different dilation rates. In our method, we use three dilated convolutions (i.e., $d_1$, $d_2$, and $d_3$) and keep the filter weight $W$ fixed. The HDC output of the input feature map $C_i^D$ can be formulated in the following:
\begin{equation}\label{eq:hdc}
%\small
\mathcal{H}(C_i^D)=\{\mathcal{D}_{r_1}(C_i^D);\mathcal{D}_{r_2}(C_i^D);\mathcal{D}_{r_3}(C_i^D);\mathcal{M}(C_i^D)\}
\end{equation}
where $\mathcal{H}$ is the concatenation of the dilated convolution results $\mathcal{D}(C_i^D)$ and dimension mapping results $\mathcal{M}(C_i^D)$. We denote the concatenated results as $H_i^D$. The dimension mapping operation $\mathcal{M}$ is a $1\times1$ convolution on the input feature maps to ensure the channel consistency with respect to the dilated convolution results, while maintaining the original feature information from the FPN.
%with respect to the dilated convolution results.
We use different dilation rates to capture the lesion responses from each pyramid feature map respectively. These fine-grained feature responses of HDC contain multiple scales of reception fields within each feature pyramid level $C_i^D$. In order to only capture the scale variation responses on the pyramid feature maps, we share weights among HDC to overcome other interferences such as rotation and deformation.

\subsection{Channel Attention}
We refine the HDC result using a squeeze-and-excitation network as shown in Fig. \ref{fig:pipeline} following \cite{hu2018squeeze}. The HDC result $H_i^D$ captures the feature responses of the potential lesions from the multi-scale perspective. For a particular lesion with a certain dimension, high feature response may reside in one of the dilated convolution scales. Therefore it is intuitive to attend the network to the subchannels of $H_i^D$.
%As the response resides in one of the dilated convolution results, we need to let the network attend to the subchannels of $H_i^\downarrow$.
We propose a channel attention module as shown in Fig. \ref{fig:pipeline}. It first squeezes $H_i^D$ by a global pooling operation and then activates the reduced feature maps by a $1\times1$ convolution layer. The channel attention can be written as:
\begin{equation}\label{eq:ch}
%\small
  \mathcal{F}_{ch}(H_i^D)=\mathcal{P}_{avg}(H_i^D)*W_{1\times1}
\end{equation}
where $\mathcal{P}_{avg}$ and $W_{1\times1}$ represent the global pooling and the convolution operation, respectively. The channel attention output $\mathcal{F}_{ch}(H_i^D)$ is a one dimensional vector re-weighting $C_i^D$. The network is learned to pay more attention to the subchannels of $H_i^D$ where the precise scale response of the lesion region resides. The reweighted feature maps from channel attention can be written as:
\begin{equation}\label{eq:caagg}
%\small
  H_i^{D ch}=\mathcal{F}_{ch}(H_i^D)\otimes H_i^D
\end{equation}
where $\otimes$ is the element-wise multiplication operation.
%where the reweighed $C_i^\downarrow$ is combined with the top-down feature map $C_i^\uparrow$ via an element-wise addition.

\subsection{Spatial Attention}
The channel attention ensures the network to focus on $H_i^{D ch}$, where the response of the scale estimation from HDC resides. To increase the network's attention to the lesion response within $H_i^{D ch}$, we propose a spatial attention module that reduces the distraction outside of the ROIs. The proposed spatial attention module first squeezes $H_i^{D ch}$ by using a max pooling operation and then performs a $3\times3$ convolution. The spatial attention activation process can be written as:
\begin{equation}\label{eq:sp}
%\small
    \mathcal{F}_{sp}(H_i^{D ch})=\mathcal{P}_{max}(H_i^{D ch})*W_{3\times3}
\end{equation}
where $\mathcal{P}_{max}$ and $W_{3\times3}$ are the max pooling and the convolution operation, respectively. The spatial attention $\mathcal{F}_{sp}(H_i^{D ch})$ is a one-channel feature map used to filer out the irrelevant information of $H_i^{D ch}$. As a result, the network will attentively focus around the lesion region. The refined output feature map can be formulated as:

%The refined feature map via spatial attention is combined with the top down $C_i^\uparrow$ to formulate the output feature map for lesion detection. The formulation can be written as:
\begin{equation}\label{eq:spagg}
%\small
  %\hat{P}_i=(\mathcal{F}_{sp}(H_i^{\downarrow ch})\otimes H_i^{\downarrow ch})\oplus C_i^\uparrow
  \hat{P}_i=\mathcal{F}_{sp}(H_i^{D ch})\otimes H_i^{D ch}
\end{equation}
where $\oplus$ is the same as that in Eq. \ref{eq:fpn}. The output feature map $\hat{P}_i$ is then used for lesion detection.

\begin{table}[t]
\centering
\caption{An ablation study with various configurations of the proposed modules. Lesion detection sensitivity is reported at different false positive (FP) rates on the DeepLesion \cite{yan2017deeplesion} test set.}
\label{Tab:ablation}
\scriptsize
\begin{tabular}{m{2.65cm}m{1.65cm}m{1.55cm}<{\centering}m{1.55cm}<{\centering}m{1.55cm}<{\centering}m{1.55cm}<{\centering}m{1.55cm}<{\centering}}
\hline
\multirow{2}{*}{Method} & \multirow{2}{*}{Backbone} & \multicolumn{5}{c}{FPs per Image}\\
\cline{3-7}
 &  & $0.5$ & $1$ & $2$ & $4$ & $8$ \\
\hline
FPN & ResNet-50 & $0.621$ & $0.728$	& $0.807$	& $0.864$	& $0.890$ \\
FPN+HDC (weights sharing) & ResNet-50 & 0.622 & 0.734 & 0.818 & 0.873 & 0.910 \\
FPN+HDC+CH (weights sharing)&  ResNet-50 & 0.645 & 0.746 & 0.820 & 0.880 & 0.911 \\
FPN+HDC+SP (weights sharing)& ResNet-50 & 0.629 & 0.743 &	0.821 & 0.881 & 0.914  \\
FPN+MSB & ResNet-50 & 0.637 & 0.748 & 0.819 & 0.871 & 0.917\\
FPN+MSB (weights sharing) & ResNet-50 &\textbf{0.670} & \textbf{0.768} & \textbf{0.837} & \textbf{0.890} & \textbf{0.920}\\
\hline
\end{tabular}
%\vspace{-4mm}
\end{table}

\section{Experiments}
We evaluate the proposed method on the large-scale benchmark dataset DeepLesion \cite{yan2017deeplesion}. It includes 32,735 lesions from 32,120 CT slices, which are captured from 4,427 patients. The lesion areas cover liver, lung nodules, bone, kidney, and other organs.
We follow the dataset configuration to split into the training, validation and test sets.
In the training process, we use ResNet50 \cite{he2016deep} as the feature extraction backbone. The initial weights from conv1 to conv5 are from the ImageNet pretrained model \cite{olga-ijcv15-imagenet} and the remaining weights are randomly initialized. We resize the CT slices to $512\times512$ pixels and concatenate three consecutive CT slides as the input to predict lesions of the central slice. The five anchor scales and three anchor ratios are set as $(8, 16, 32, 64, 128)$, $\{1:2, 1:1, 2:1\}$ respectively at each level while training RPN. The learning rate is set as 0.01 and the learning process is around 10 epochs.

%\subsection{Experimental Setup}
%We trained our framework by using pytorch-0.4.1 on NVIDIA Tesla $P40$ GPU. We choose ResNet50\cite{he2016deep} as backbone, and initialize weights of conv1-conv5 with an ImageNet pretrained model. All other weights are initialized with Gaussian distribution with $mean = 0$ and $std =0.01$. The five anchor scales and three anchor ratios are set as $(8, 16, 32, 64, 128)$, $\{1:2, 1:1, 2:1\}$ respectively at each level while training RPN. For the purpose of encode three-dimensional information of date, we aggregated three neighboring slice into a three channel image during all the experiments. We resized all CT slice into $512 \times 512$ and set batch size as $16$ (i.e. $16 \times 3 = 48$ slice were used at one batch) during training. We applied stochastic gradient descent(SGD) as optimizer and set base learning rate as $0.01$, then reduced it by CosineAnnealingLR at each epoch. The whole training stopped at 12 epoch. IoU was used as the overlap computation criterion during all experiments.

\subsection{Ablation Study}\label{sec:ablation}
The proposed network consists of four major components. They are FPN, HDC, CH (channel attention), and SP (spatial attention). To evaluate the effectiveness of each module and weights sharing, we ablatively study on the DeepLesion dataset. The evaluation metric is the average sensitivity values at different false positives rates of the whole test set. The evaluation configuration is shown in Table \ref{Tab:ablation}.
%On the first row, we show the performance by only using FPN. On the second row, we show the detection performance when FPN is combined with HDC. On the third row and fourth row, we show the performance when FPN+HDC is combined with CH and SP, respectively. On the last row, we show the performance of the proposed method where MSB consists of HDC, CH, and SP.
The comparisons among different configurations demonstrate that the proposed MSB achieves highest sensitivity under different false positives rates.

\subsection{Comparisons with State-of-the-art}
We compare the proposed method with state-of-the-art approaches including 3DCE \cite{yan20183d}, Faster R-CNN \cite{ren-pami16-faster} and FPN \cite{lin-iccv17-fpn}.
%yb: clear
Yan et al. \cite{yan20183d} sent multiple slices into the 2D detection network (i.e., Faster R-CNN \cite{ren-pami16-faster}) to generate feature maps separately, and then aggregated them to incorporate 3D context information for final prediction.
%yb: you can remove this sentence.
We note that the results of 3DCE \cite{yan20183d} are the only available results reported on this dataset.
%To the best of our acknowledge, results of 3DCE claimed in \cite{yan20183d} is the only published results on this dataset}.
%
We perform the evaluation from two perspectives. The first one is to compute the sensitivity values at different false positives rates as illustrated in Section \ref{sec:ablation}. It reflects the averaged performance of each method for test set. The other one is to compute the sensitivity values generated based on different sizes of lesions. It reflects how effective each method is to detect lesions at different scales.

\begin{table}[t]
\centering
\caption{Comparison of the proposed method (FPN + MSB) with state-of-the-art methods on the DeepLesion \cite{yan2017deeplesion} test set. Lesion detection sensitivity values are reported at different false positive (FP) rates.}
\label{Tab:Sensitivity on official test set}
\scriptsize
\begin{tabular}{m{2.5cm}m{1.5cm}m{2.3cm}<{\centering}m{1.15cm}<{\centering}m{1.15cm}<{\centering}m{1.15cm}<{\centering}m{1.15cm}<{\centering}m{1.15cm}<{\centering}}
\hline
\multirow{2}{*}{Method} & \multirow{2}{*}{Backbone} & \multirow{2}{*}{Number of slices} & \multicolumn{5}{c}{FPs per Image} \\
\cline{4-8}
&  & &  0.5 & 1 & 2 & 4 & 8 \\
\hline
\multirow{3}{*}{3DCE \cite{yan20183d}} & VGG-16 & 3 & 0.569 & 0.673 &  0.756 & 0.816 & 0.858 \\
 & VGG-16 & 9 & 0.593 & 0.707 &  0.791 & 0.843 & 0.878 \\
 & VGG-16 & 27 & 0.625 & 0.737 & 0.807 & 0.857 & 0.891 \\
Faster R-CNN \cite{ren-pami16-faster} & ResNet-50 & 3 & 0.560 & 0.677 & 0.763 & 0.832 & 0.867 \\
FPN \cite{lin-iccv17-fpn} & ResNet-50 & 3 & $0.621$ & $0.728$	& $0.807$	& $0.864$	& $0.890$ \\
FPN+MSB (weights sharing) & ResNet-50 & 3 & \textbf{0.670} & \textbf{0.768} & \textbf{0.837} & \textbf{0.890} & \textbf{0.920}\\
\hline
\end{tabular}
%\vspace{-4mm}
\end{table}

Table \ref{Tab:Sensitivity on official test set} shows the evaluation results. It demonstrates that the proposed method performs superiorly against existing methods.
We note that there are different numbers of CT slices used as input for 3DCE to produce different sensitivity values.
The result shows that sensitivity value increases when more CT slices are taken as input. As these CT slices are captured on the same organ of the patient, using more slices will provide sufficient information to the network to detect.
Nevertheless, we show that the proposed method achieves higher sensitivity values when using only three slices as input.

\begin{table}[t]
\centering
\caption{Sensitivity values at four false positives per image on five test subsets categorized by different lesion size.}
\label{Tab:Sensitivity on different scales}
\scriptsize
\begin{tabular}{m{2.5cm}m{1.5cm}m{2.3cm}<{\centering}m{1.15cm}<{\centering}m{1.15cm}<{\centering}m{1.15cm}<{\centering}m{1.15cm}<{\centering}m{1.15cm}<{\centering}}
\hline
\multirow{2}{*}{Method} & \multirow{2}{*}{Backbone} & \multirow{2}{*}{Number of slices} & \multicolumn{5}{c}{Lesion Diameters(mm)}\\
\cline{4-8}
 & & & $< 10$ & $10 \sim 30$ & $30 \sim 60$ & $60 \sim 100$ & $> 100$ \\
\hline
3DCE\cite{yan20183d} & VGG-16 & 27 & 0.78 & 0.86 & \multicolumn {3}{c}{0.84} \\
Faster R-CNN \cite{ren-pami16-faster} & ResNet-50 & 3 & 0.77 & 0.86 & 0.81 & 0.88 & 0.72 \\
FPN \cite{lin-iccv17-fpn} & ResNet-50 & 3 & 0.83 & 0.88	& 0.82	& 0.91	& 0.77 \\
FPN+HDC (weights sharing)& ResNet-50 & 3 & 0.85 & 0.89	& \textbf{0.88}	& \textbf{0.93}	& 0.79 \\
FPN+MSB (weights sharing) & ResNet-50 & 3 & \textbf{0.86} & \textbf{0.91} & 0.86 & \textbf{0.93} & \textbf{0.86} \\
\hline
\end{tabular}
%\vspace{-4mm}
\end{table}

To evaluate how the proposed method performs when detecting different size of lesions, we divide the test set into five categories. Each category consists of lesions in a fixed range of size and the range does not overlap with each other. Table \ref{Tab:Sensitivity on different scales} shows the evaluation results. The proposed method shows better performance to detect lesions in different sizes. Meanwhile, the sensitivity values of the proposed method exceed those of existing methods more when the size of the testing lesions becomes extremely large or small (i.e., the diameters of the lesions are above 100 mm or below 10 mm). It indicates that the proposed method is more effective to detect extreme scales of the input lesions.

\section{Conclusion}
We proposed a multi-scale booster (MSB) to detect lesion in large scale variations. We use FPN to decompose the feature map response into several coarse-grained pyramid levels. Within each level, we increase the network awareness of the scale variations by using HDC. The HDC offers fine-grained scale estimations to effectively capture the scale responses. To effectively select meaningful responses, we proposed a cascaded attention module consists of channel and spatial attentions. Evaluations on the DeepLesion benchmark indicated the effectiveness of the proposed method to detect lesions at vastly different scales.

%
% ---- Bibliography ----
%
% BibTeX users should specify bibliography style 'splncs04'.
% References will then be sorted and formatted in the correct style.
%
\bibliographystyle{splncs04}
\bibliography{ref}
%
%\begin{thebibliography}{8}
%\bibitem{ref_article1}
%Author, F.: Article title. Journal \textbf{2}(5), 99--110 (2016)

%\bibitem{ref_lncs1}
%Author, F., Author, S.: Title of a proceedings paper. In: Editor,
%F., Editor, S. (eds.) CONFERENCE 2016, LNCS, vol. 9999, pp. 1--13.
%Springer, Heidelberg (2016). \doi{10.10007/1234567890}

%\bibitem{ref_book1}
%Author, F., Author, S., Author, T.: Book title. 2nd edn. Publisher,
%Location (1999)

%\bibitem{ref_proc1}
%Author, A.-B.: Contribution title. In: 9th International Proceedings
%on Proceedings, pp. 1--2. Publisher, Location (2010)

%\bibitem{ref_url1}
%LNCS Homepage, \url{http://www.springer.com/lncs}. Last accessed 4
%Oct 2017
%\end{thebibliography}
\end{document}